\documentclass[letterpaper]{article} % DO NOT CHANGE THIS
\usepackage[]{aaai23}  % DO NOT CHANGE THIS
\usepackage{times}  % DO NOT CHANGE THIS
\usepackage{helvet}  % DO NOT CHANGE THIS
\usepackage{courier}  % DO NOT CHANGE THIS
\usepackage[hyphens]{url}  % DO NOT CHANGE THIS
\usepackage{graphicx} % DO NOT CHANGE THIS
\usepackage{natbib}  % DO NOT CHANGE THIS AND DO NOT ADD ANY OPTIONS TO IT
\usepackage{caption} % DO NOT CHANGE THIS AND DO NOT ADD ANY OPTIONS TO IT
\usepackage{algorithm}
\usepackage{algorithmic}

\usepackage{amsmath}
\usepackage{multirow}
\usepackage{subfigure}
\usepackage{booktabs}
\usepackage{newfloat}
\usepackage{listings}
\DeclareCaptionStyle{ruled}{labelfont=normalfont,labelsep=colon,strut=off} % DO NOT CHANGE THIS
\floatstyle{ruled}
\newfloat{listing}{tb}{lst}{}
\floatname{listing}{Listing}
\nocopyright
\title{Feature Weaken: Vicinal Data Augmentation for Classification}
\author{
    Songhao Jiang\textsuperscript{1,3} \hspace{2em} 	Yan Chu\textsuperscript{2}\thanks{Corresponding Author  }  \hspace{2em}  Tianxing Ma\textsuperscript{1,3}\hspace{2em} 
Tianning Zang\textsuperscript{1,3}\thanks{Corresponding Author  }\hspace{2em} 
}
\affiliations{
    \textsuperscript{1}Institute of Infomation Engineering, Chinese Academy of Sciences \\
	  \textsuperscript{2}Harbin Engineering University\\
	  \textsuperscript{3}School of Cyber Security, University of Chinese Academy of Sciences\\
            jiangsonghao@iie.ac.cn
	\hspace{1em} chuyan@hrbeu.edu.cn
        \hspace{1em} matianxing2000@163.com
        \hspace{1em} zangtianning@iie.ac.cn
}

\usepackage{bibentry}
\begin{document}

\maketitle

\begin{abstract}
Deep learning usually relies on training large-scale data samples to achieve better performance. However, over-fitting based on training data always remains a problem. Scholars have proposed various strategies, such as feature dropping and feature mixing, to improve the generalization continuously. For the same purpose, we subversively propose a novel training method, Feature Weaken, which can be regarded as a data augmentation method. Feature Weaken constructs the vicinal data distribution with the same cosine similarity for model training by weakening features of the original samples. In especially, Feature Weaken changes the spatial distribution of samples, adjusts sample boundaries, and reduces the gradient optimization value of back-propagation. This work can not only improve the classification performance and generalization of the model, but also stabilize the model training and accelerate the model convergence. We conduct extensive experiments on classical deep convolution neural models with five common image classification datasets and the Bert model with four common text classification datasets. Compared with the classical models or the generalization improvement methods, such as Dropout, Mixup, Cutout, and CutMix, Feature Weaken shows good compatibility and performance. We also use adversarial samples to perform the robustness experiments, and the results show that Feature Weaken is effective in improving the robustness of the model.
\end{abstract}

\section{Introduction}
 
The advantages of deep learning usually have a linear relationship with model parameters and the scale of training samples. Large-parameter models and large-scale training samples have become a trend in deep learning model research \cite{bommasani2021opportunities}. For example, GPT-3 has 175 billion parameters and is trained on broad data at scale \cite{brown2020language}. Deep learning models usually adopt the training method for minimizing the risk of the experience distribution of training samples, which is called the empirical risk minimization (ERM) principle \cite{vapnik1998statistical}. In ERM mode, more training is conducted via memorization of data by model. However, a typical problem of this method is poor generalization, which can easily lead to over-fitting. And it often fails to achieve the same good performance as the training samples in the approximate distribution of the test samples.  

To solve the problem of model generalization and over-fitting, scholars propose multiple solutions. Feature dropping methods, such as Dropout \cite{hinton2012improving,srivastava2014dropout} and Cutout \cite{devries2017improved}, improve the model generalization ability during training by dropping some features and preventing co-adaptation of feature detectors. Normalization methods use standard deviations and variances to uniformly expand or shrink a certain range of sample features to improve the generalization and training speed \cite{ioffe2015batch,wu2018group,ba2016layer}. Many scholars further study other training techniques to improve model generalization, such as classical regularization, label smoothing, and weight decay. However, such methods usually do not change the training pattern of ERM.

\begin{figure}[t]
	\centering
	\subfigure[Original]{
		\includegraphics[width=0.3\linewidth]{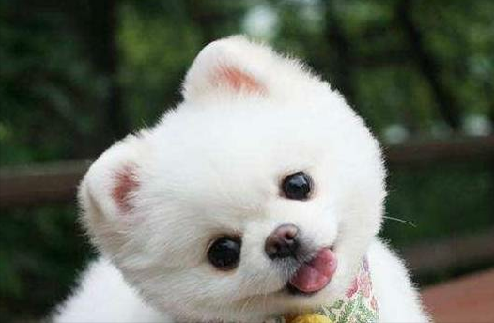}}
	\subfigure[Mixup]{
		\includegraphics[width=0.3\linewidth]{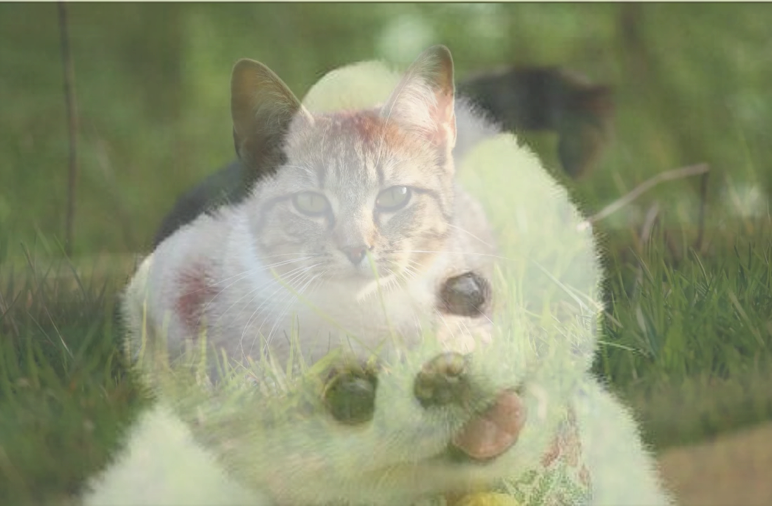}}
	\subfigure[Cutout]{
		\includegraphics[width=0.3\linewidth]{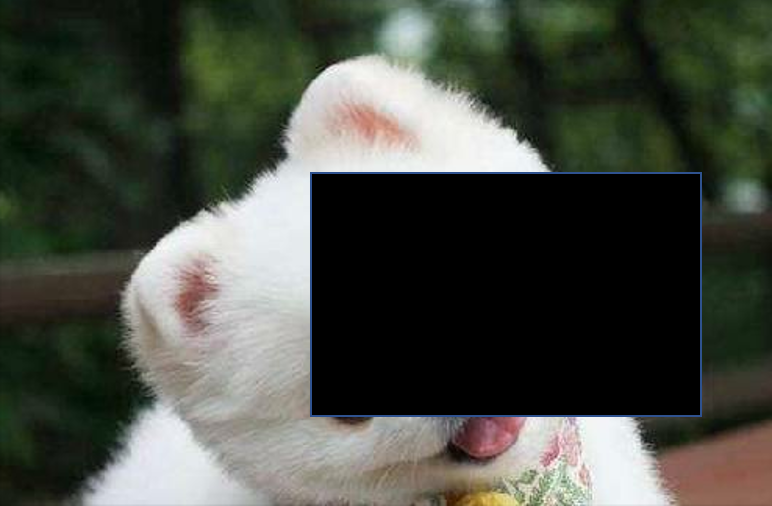}}
	\subfigure[CutMix]{
	\includegraphics[width=0.3\linewidth]{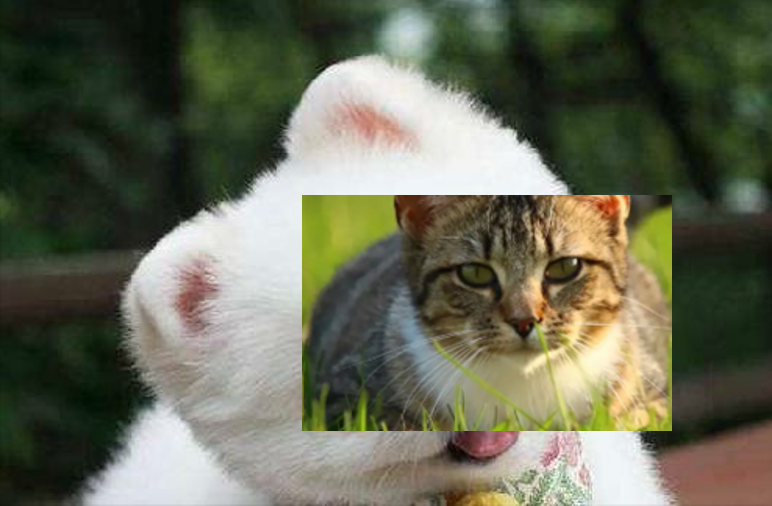}}
	\subfigure[Feature Weaken]{
		\includegraphics[width=0.3\linewidth]{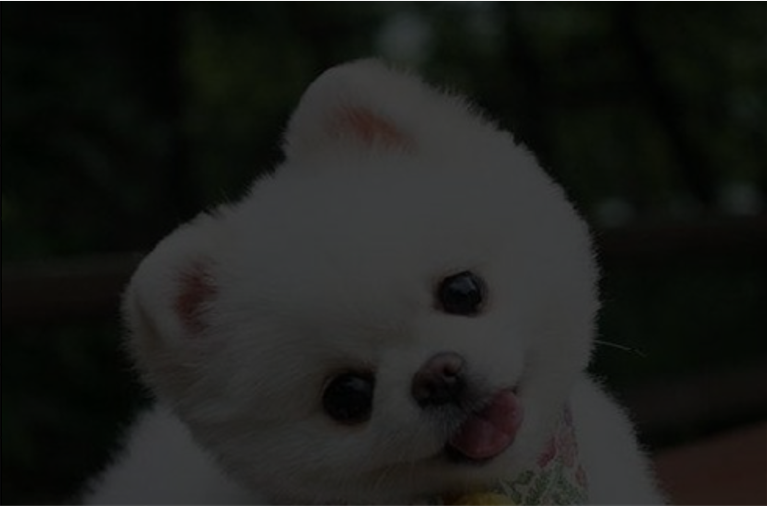}}
	\caption{Overview of the results of Mixup, Cutout, CutMix, and Feature Weaken on example images.}
	\label{Fig1}
\end{figure}

The data augmentation method can increase the scale of training samples by generating some new samples. Instead of using the empirical distribution of original samples, the vicinal distribution is used for training to improve the generalization and solve the over-fitting problems. Therefore, the ERM principle is changed to a vicinal risk minimization (VRM) principle \cite{chapelle2000vicinal,zhang2017mixup}. But, standard data augmentation methods often rely on expert knowledge to change data according to the characteristics of different data and transform the original distribution of samples, such as flipping and distorting changes in images or easy data augmentation methods (EDA) in texts \cite{wei2019eda,guo2021have}.  
Feature mixing is another kind of data augmentation, which can realize vicinal changes of samples independent of the characteristics of samples, such as Mixup \cite{zhang2017mixup} and CutMix \cite{yun2019cutmix}.

In this paper, we consider improving generalization. Unlike previous methods of feature dropping and feature mixing, we propose a novel training method of vicinal risk minimization, \textbf{Feature Weaken}, which also realizes the vicinal distribution change of samples independently of the samples characteristics and expert knowledge. Feature Weaken can also be regarded as a data augmentation method. It mainly reduces and weakens the vectors in the embedding-level or hidden-level to transform the spatial distribution of the original samples without changing the spatial angle of the samples in the same coordinate system. When applying Feature Weaken on the embedding-level, a representative comparison of our method with Mixup, CutMix, and Cutout is shown in Figure \ref{Fig1}. Feature Weaken can also coexist with other data augmentation methods.  When Feature Weaken combines with Mixup or CutMix, it can better come into the vicinal changes of samples as well as realize feature fusion learning among samples.

We verify the effectiveness of Feature Weaken through extensive experiments. In image classification, we use classical models such as ResNet \cite{he2016deep} and DenseNet \cite{huang2017densely}. We conduct experiments in MNIST \cite{lecun1998gradient}, Fashion-MNIST \cite{xiao2017fashion}, STL-10 \cite{coates2011analysis}, CIFAR-10, and CIFAR-100 \cite{krizhevsky2009learning} image classification datasets. Compared with Mixup, Dropout, Cutout, CutMix, and other methods, Feature Weaken achieves an absolutely leading performance. In text classification, we use Bert \cite{devlin2019bert} as the backbone model. Compared with advanced text data augmentation methods SenMixup \cite{guo2019augmenting}, TMix \cite{chen2020mixtext} and SSMIX \cite{yoon2021ssmix} in SST \cite{socher2013recursive}, and TREC \cite{li2002learning} datasets, it exhibits a top performance. In addition, FGSM and I-FSGM \cite{goodfellow2014explaining} are used to generate adversarial samples, and the results show that Feature Weaken can improve the robustness of the model.

\section{Related Work}

\paragraph{Feature Drop:} Feature Drop is a common regularization strategy. According to the different positions of feature dropping, it can be roughly divided into two categories. One is the feature dropping of the hidden-level to achieve the neuron to focus on the general features rather than the part of features. Dropout \cite{hinton2012improving,srivastava2014dropout}, DropBlock \cite{ghiasi2018dropblock} and other variants \cite{,ba2013adaptive,wan2013regularization,park2016analysis,keshari2019guided,choe2019attention,pham2021autodropout} selected the feature elements, feature regions, channels, neurons, neural network paths of the hidden layer to drop during the model training process to achieve a better model generalization performance. Another is the feature dropping of original samples to augment the sample data and improve the generalization of the model. Cutout \cite{devries2017improved}, Random Erasing \cite{zhong2020random}, HaS\cite{singh2018hide}, and other methods \cite{chen2020gridmask,gong2021keepaugment} can achieve the dropping of original sample features by randomly blocking or inserting noise. Nevertheless, these methods may cause significant sample areas to be removed, leading to insufficient features for deep learning models. Therefore, the feature dropping method affects the learning effect of the model and weakens the joint effect of network nodes and features. Unlike Feature Drop, Feature Weaken preserves the global features to ensure that the original information is not lost. 

\paragraph{Feature Mix:}  Feature Mix is another regularization method that has been paid attention to by scholars in recent years. It mainly achieves sample data augmentation through different levels of feature mixing to improve the generalization of the model. Applying data augmentation technology can overcome over-fitting and improve generalization effectively. For example, in NLP, some scholars have augmented the text data through synonym replacement, random swap, random insertion, and random deletion to improve the model generalization \cite{wei2019eda,guo2021have}. However, these methods require more expert knowledge for guidance. Mixup and its variants realized the training method beyond empirical risk minimization, and utilized linear interpolation theory to synthesize data and labels \cite{zhang2017mixup,guo2019Mixup,verma2019manifold,guo2020nonlinear,sawhney2022dmix}. CutMix and its variants used samples of other categories to replace original regions \cite{yun2019cutmix,faramarzi2022patchup,walawalkar2020attentive,kim2020puzzle,chen2022transmix}.  Moreover, due to the discreteness of text features, text mixing augmentation methods mostly occurred in the representative layer and hidden layer \cite{guo2019Mixup,chen2020mixtext}. While, SSMIX \cite{yoon2021ssmix}, TreeMix \cite{zhang2022treemix} and other methods \cite{shi2021substructure,kim2022alp} realized the mixing of input-level samples by using text structure. Although the feature mixing was proved to be a VRM pattern beyond ERM  \cite{zhang2017mixup}.  Feature Weaken uses weakening operation to transform the vicinal distribution of the original samples, which is completely different from feature mixing or other data augmentation methods. As well as our method adjusts the intensity of feature space, does not need huge computing costs, and can be used in combination with other methods.

\paragraph{Other regularization  methods:} To improve the generalization of deep learning algorithms, other regularization methods were proposed. Normalization methods used standard deviations and variances to uniformly expand or shrink a certain range of training samples or model weights \cite{ioffe2015batch,wu2018group,ba2016layer}. These methods concentrated the weight representation in a certain range and made the optimization landscape significantly smoother \cite{santurkar2018does,bjorck2018understanding}. As a result, the model generalization and training speed were improved. In addition, many scholars also studied other training techniques to improve model generalization, such as classical regularization, label smoothing, weight decay, and early stopping \cite{nowlan1992simplifying,szegedy2016rethinking,zhang2021delving,muller2019does,lienen2021label,guo2021label,loshchilov2017decoupled,zhang2018three,krogh1991simple,bai2021understanding,heckel2020early}. These methods usually improve the generalization of the model by increasing the training difficulty or limiting the training loss. Although Feature Weaken also affects gradient optimization in model training, these methods do not change the features of samples. As well as, these methods did not get rid of the training pattern of ERM.

% In the process of training, a large number of calculations need to be carried out on data, and improving the efficiency of network training has been widely concerned. The number of iterations of the network is not as large as possible. Early stopping makes the model stop optimization at an appropriate time to prevent over-fitting. Nevertheless, how to choose an appropriate stopping point is still being explored (Bai et al.2021; Heckel and Yilmaz.2021). The size of batch size will affect the convergence speed of the network and the efficiency of gradient descent (He, Liu, and Tao. 2019). In classification tasks, labels are often represented by one hot vectors. If the model only pays attention to the loss of correct label position during training, it will lead to training deviation. Label smoothing as a regularization method can effectively avoid this phenomenon (Szegedy et al.2016; Guo et al.2021).

\section{Methodology}

\subsection{Feature Weaken}

\begin{figure}[t]
	\centering
	\includegraphics[width=0.9\linewidth]{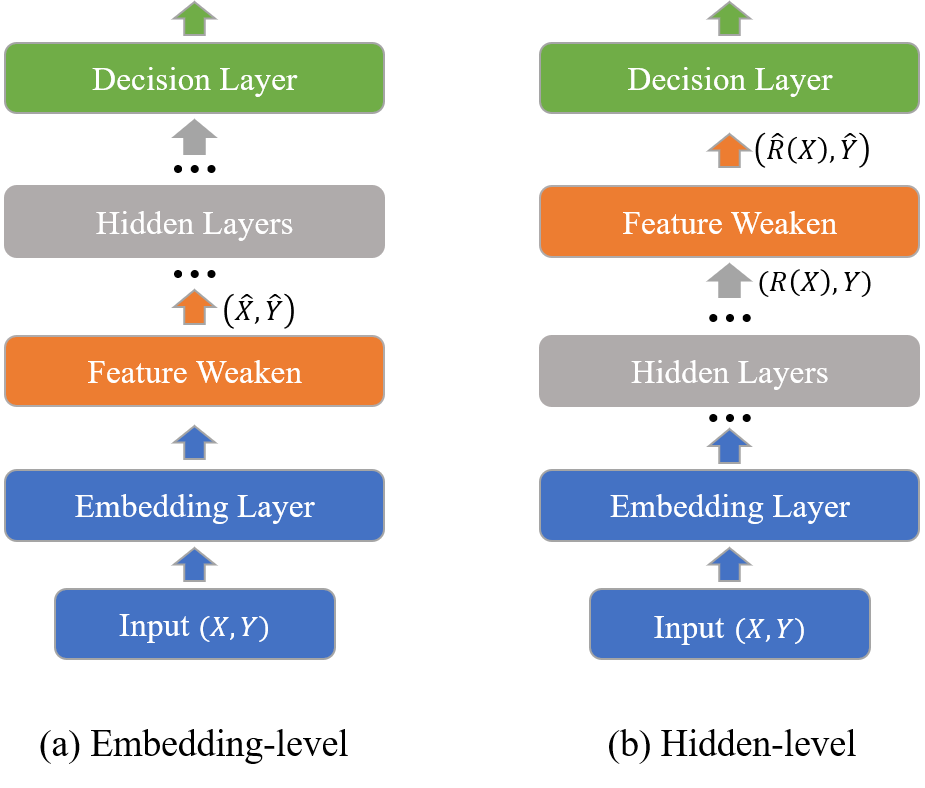}
% 	\subfigure[Ebedding-level]{
% 		\includegraphics[width=0.4\linewidth]{el.png}}
% 	\subfigure[Hidden-level]{
% 		\includegraphics[width=0.4\linewidth]{hl.png}}
	\caption{The overview of the deep model with Feature Weaken. (a) shows Feature Weaken on the embedding-level, and (b) shows Feature Weaken on the hidden-level.}
	\label{Fig0}
\end{figure}

As shown in Figure \ref{Fig0}, we put forward two methods, one is embedding-level Feature Weaken (FW-el), and the other is hidden-level Feature Weaken (FW-hl). In the embedding-level Feature Weaken mode, we use $ X\in\mathbf{R}^{W \times H \times C}$ to represent the original training image/sample, and $ Y $ to represent the label of the image/sample. Feature Weaken aims to generate a vicinal distribution sample $ (\hat{X},\hat{Y})  $ of $ (X,Y) $ by weakening the feature. We define the combining operation as:
\begin{align}
	\hat{X} = (1-Ws) \cdot X,
	\hat{Y} = Y.
\end{align}
Where $Ws\in(0,1)$ represents the parameter of Feature Weaken and refers to the weaken strength of $X$. The cosine similarity of $X$ and $\hat{X}$ remains the same and does not affect the label of the sample. Therefore, we assign the value of $Y$  to $ \hat{Y}$. The new samples $\hat{X}$ and the label $\hat{Y}$ are used for model training with the original loss function.

For hidden-level Feature Weaken, we consider weakening the representation features of samples before the decision layer. Because the representation tensors are extracted by the deep model, and the representation tensors of samples are closer to the decision function. We think it is more conducive to model training. Therefore, for the representation features to weaken, the representation change of $ (X, Y) $ as defined in the following equation:
\begin{align}
	\hat{R}(X) = (1-Ws) \cdot R(X),
	\hat{Y} = Y.
\end{align}
Where $ R (.) $ indicates the sample tensors extracted by the deep model, and $ Ws $ indicates the weaken strength. The $(\hat{R}(X),\hat{Y})$ is the weaken data of $(R(X),Y)$.
\subsection{The Change of The Loss Optimization}
For supervised training, we need to consider that the function $ f $ of a model can adequately describe the relationship between sample $ X $ and label $ Y $ in the distribution $ P(X, Y) $. 
According to ERM \cite{vapnik1998statistical}, it is generally considered that the known distribution of the training data set $ D $ is regarded as an empirical distribution approximately equivalent to $ P $. The training function of a common ERM is shown as follows:
\begin{align}
	E_{(X,Y) \sim P_D } = \frac{1}{n}\sum_{i}^{n} loss(f(x_i),y_i).
\end{align}
The model is trained by minimizing risk $ E $. For Feature Weaken, we try to use the vicinal distribution $P_{FW}(\hat{X},\hat{Y}|X,Y)$ of weakened samples $ (\hat{X},\hat{Y}|X,Y)  $, replacing the original empirical distribution $ P_D (X,Y) $, to augment the training sample data and improve the generalization ability. Therefore, Feature Weaken can be seen as a data augmentation method. The training function of Feature Weaken is changed as follows:
\begin{align}
	E_{(\hat{X},\hat{Y}) \sim P_{FW} } &= \frac{1}{n}\sum_{i}^{n} loss(f(\hat{x_i}),\hat{y_i})\\
	&= \frac{1}{n}\sum_{i}^{n} loss(f((1-Ws) \cdot x_i),y_i).
\end{align}
So, when the method of gradient descent is used for parameter optimization of back-propagation, the gradient value $\nabla$ of the vicinal sample $ (\hat{X},\hat{Y}) $ changes as follows:
\begin{gather}
	\nabla = \frac{\partial loss}{\partial \theta} = {loss}'(f_{\theta}(\hat{X}),\hat{Y})\cdot{f_\theta}'(\hat{X}),\\
	{f_\theta}'(\hat{X}) = \hat{X},\\
	\nabla = {loss}'\cdot \hat{X} = (1-Ws) \cdot {loss}'\cdot X.
\end{gather}
Compared with the gradient of loss function of the original sample, the gradient value is reduced due to the $ Ws $ of Feature Weaken. We can also regard this change as a decay of the training gradient, which has a similar effect to the learning rate or weight decay commonly seen in SGD training. Therefore, Feature Weaken can improve the model generalization by changing the gradient value of loss optimization. However, Feature Weaken is gradient decay derived from augmented training data, unlike the learning rate or weight decay, which just changes the updated gradient.

\subsection{The Change of Space Vector}

\begin{figure}[t]
	\centering
	
%	\rule[-.5cm]{0cm}{4cm}
	\includegraphics[width=0.9\linewidth]{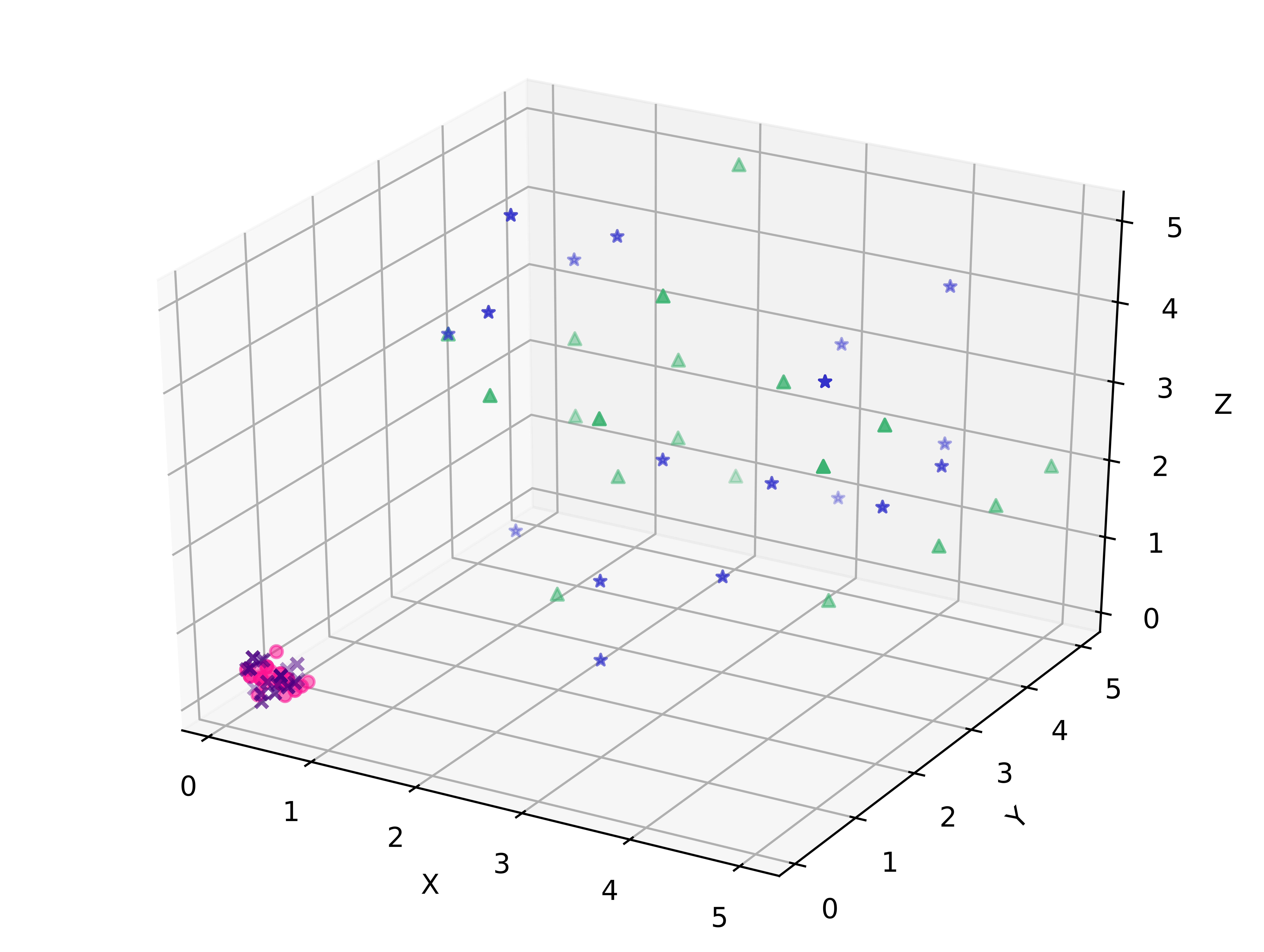}
%	\rule[-.5cm]{4cm}{0cm}
	
	\caption{The 3-dimensional scatters of original samples and weakened samples. The $\triangle$ and $\star$ indicate the original samples. The $\times$ and $\bullet$ indicate the weakened samples.}
	\label{Fig2}
\end{figure}

Feature Weaken does not change the feature dimension or representation space of the original sample, but moves and weakens the original sample vector or representation tensor towards the coordinate axis origin in the same feature space. As shown in Figure \ref{Fig2}, the original sample features and weakened features are mapped to a 3-dimensional coordinate system. In the same dimensional space, the original samples are relatively scattered. The boundary spacing of original samples is wider and easier to train. Feature Weaken makes the features more compact and makes it harder to distinguish between the boundary of the model training.

In addition, the angle between the space vector of the weakened samples with the coordinate axis remains the same as the original samples. We calculate it based on cosine similarity $ cosine (X,(1-Ws) \cdot X) $, and cosine similarity is 1. Feature Weaken can be regarded as moving along the vector axis, that formed by the center of the original sample and the coordinates axis origin, towards the origin of the coordinates. The weakened sample vectors do not change the spatial angle of the original sample vectors. And the boundary spacing of the sample distribution is simultaneously scaled down. In model training, since the cosine similarity of the two space vectors is consistent, the segmentation boundary learned by the model can realize the segmentation of weakened samples and original samples. Therefore, Feature Weaken can obtain the vicinal distribution, and the vicinal risk minimization training can be well applied to the classification decision of original samples.

\section{Experiments}
In this section, we evaluate Feature Weaken for its capability to improve the performance and generalization of models on multiple tasks. We first study the effect of Feature Weaken on image classification (Section 4.1) and text classification (Section 4.2). Next, we show that Feature Weaken can improve the model robustness (Section 4.3). We analyze the influence of Feature Weaken parameters in Section 4.4. All experiments run on multiple NVIDIA TITAN XP GPUs with PyTorch.

\subsection{Image Classification Experiments}
\paragraph{Image datasets:} In image classification experiments, we use five common datasets and utilize the \emph{torchvision.datasets}\footnote{\url{https://pytorch.org/vision/stable/index.html}} tool to download and call. They are as follows:
\begin{itemize}
\item \textbf{MNIST} 
\cite{lecun1998gradient} 
consists of pictures of handwritten numerals, of which there are 10 categories corresponding to the Arabic numerals 0 to 9. Among them, there are 60,000 images in the training set and 10,000 images in the test set.
\item \textbf{Fashion-MNIST} 
\cite{xiao2017fashion} 
is no longer an abstract symbol but a more concrete human clothing. There are 10 categories: T-shirt/top, Trouser, Pullover, Dress, Coat, Sandal, Shirt, Sneaker, Bag, and Ankle boot. It also has 60,000 images in the training set and 10,000 images in the test set.
\item \textbf{STL-10} 
\cite{coates2011analysis} 
consists of 113,000 colour images, but it is composed of 100,000 unlabeled images. For labeled images, the training set contains 5,000 images, while the test set consists of 8,000 images. All training and test set images belong to 10 categories, such as cat, dog, or plane. STL-10 is mainly used for testing semi-supervised learning algorithms. To observe Feature Weaken performs when applied to fewer data of higher resolution images, we only use the labeled training set.
\item \textbf{CIFAR-10 and CIFAR-100} are the CIFAR datasets 
\cite{krizhevsky2009learning} 
and contain 60,000 colour images. CIFAR-10 has 10 distinct categories, such as frog, truck, or boat. CIFAR-100 contains 100 categories. Since some of the classes are visually very similar examples, CIFAR-100 requires a finer identification than CIFAR-10. Each dataset is split into a training set with 50,000 images and a test set with 10,000 images.
\end{itemize}
\paragraph{Feature Weaken for various models:}
\begin{table}[t]
	
	\centering
	\begin{tabular}{cccc}
		\hline
		%\multicolumn{2}{c}{Part}                   \\
		%\chline
		Model    &MNIST  & Fashion-MNIST & STL-10  \\
		\hline
		ResNet-18  &  99.69&	\textbf{94.16}&	60.93
		       \\
		+ FW-hl  &   \textbf{99.71}&	94.15&	\textbf{71.39}
		        \\
		\hline
		ResNet-50  &  99.65&	93.57&	67.96
		        \\
		+ FW-hl  &   \textbf{99.67}&	\textbf{94.13}&	\textbf{68.26}
		        \\
		\hline
		ResNet-101  & 99.63&	93.72&	66.50
		        \\
		+ FW-hl  &   \textbf{99.66}&	\textbf{93.95}& 	\textbf{70.50}
		         \\
		\hline
		DenseNet-121 &  99.60&	93.01&	68.10
		         \\
		+ FW-hl  &   \textbf{99.61}&	\textbf{93.10}&	\textbf{72.86}
		       \\
		\hline
	\end{tabular}
	\caption{Feature Weaken for various models.}
	\label{table1}
\end{table}
\begin{table*}[ht]

	\centering
	\begin{tabular}{cccccccc}
		\hline
		%\multicolumn{2}{c}{Part}                   \\
		%\chline
		\multirow{2}{*}{Model}    &\multirow{2}{*}{MNIST}  & \multicolumn{2}{c}{CIFAR-10}  & \multicolumn{2}{c}{CIFAR-100}  &  \multicolumn{2}{c}{STL-10}\\
		\cmidrule(r){3-4} \cmidrule(r){5-6} \cmidrule(r){7-8}
% 		& &\multicolumn{2}{c|}{(200 epochs)}&\multicolumn{2}{c|}{(200 epochs)}&\multicolumn{2}{c}{(50 epochs)}\\
		&  &Top-1&Top-5&Top-1&Top-5&Top-1&Top-5\\
		\hline
		ResNet-18 (baseline)  &  99.66&   88.76& 99.53& 63.27& 86.5&   60.93& 96.4    \\
		+ Dropout (0.5)  &+0.0& +0.04& -0.01	&	+0.52& +0.66&		+5.58&+0.89   \\
		     
		+ Mixup ($\alpha=0.4$)  &  +0.03&   +0.66& -0.05& -0.24& -0.24& +2.66 &  -0.72     \\
		+ Cutout ($Patch Length = 16$) & +0.02&   +1.54& +0.15& +1.77& +1.95& +6.16 &  +0.86   \\
		+ CutMix ($\alpha=1.0$) & -0.03&   +3.06& +0.16& +6.69& +5.09& +4.97 &  +0.46     \\
% 		+Attentive CutMix* & -& +0.18&-&+3.89&-&-&-\\
		\hline
		+ FW-hl ($Ws=0.8$) &  +0.02&   +1.24&  +0.17&  +2.70& +2.29&  +11.46& +1.50    \\
		+ Droupout (0.5) + FW-hl ($Ws=0.8$) &  +0.03 &    +1.47&	+0.16&	+3.62&	+2.79&  +11.01& \textbf{+2.10}\\
		+ Mixup ($\alpha=0.4$) + FW-hl ($Ws=0.8$)  &   \textbf{+0.06}&   +1.45&   +0.01&  +2.30&   +1.74 & +11.52 &+0.91   \\
		+ Cutout ($Patch Length = 16$) + FW-hl ($Ws=0.8$)  &   \textbf{+0.06}&   +2.40&  +0.21&    +4.49& +3.76& +13.17 &+1.83  \\
		+ CutMix ($\alpha=1.0$) + FW-hl ($Ws=0.8$)  &   -0.02&   \textbf{+3.55}&    \textbf{+0.24}&   \textbf{+6.95}& \textbf{+5.46}& \textbf{+14.17}& +1.73 \\
		\hline
	\end{tabular}
	\caption{Comparison against other methods on  MNIST, CIFAR, and STL-10. We show the positive and negative gains of the validation accuracy (\%) of different methods compared with the baseline. }
% 	And we include results from \cite{walawalkar2020attentive}*.}
	\label{table2}
\end{table*}

% \begin{table}[ht]

% 	\centering
% 	\begin{tabular}{c|c}
% 		\hline
% 		%\multicolumn{2}{c}{Part}                   \\
% 		%\chline
% 		Model    & CIFAR-10   \\
% 		\hline
% 		WideResNet-28*  &  4.24 $\pm$ 0.14 \\
% 		+ ManifoldMixup*($\alpha = 1.5$)  &    3.27 $\pm$ 0.35\\
% 		+ DropBlock* &   4.18 $\pm$0.07\\
% 		+ Puzzle Mix* &   2.56 $\pm$ 0.07\\
% 		+ PatchUp* &  2.53 $\pm$ 0.07\\
% 		+ PatchUp + FW-hl($Ws=0.5$) &  \textbf{2.518 $\pm$ 0.07}\\

% 		\hline
% 	\end{tabular}
% 	\caption{The error rates (\%) on CIFAR-10.The best results are highlighted in bold. And we include results from \cite{faramarzi2022patchup}*.}
% 	\label{tablea}
% \end{table}
We use ResNet-18, ResNet-50, ResNet-101 and DenseNet-121 for experiments on MNIST, Fashion-MNIST and STL-10. For ResNet, we reprocess the code based on Cutout \cite{devries2017improved}. For DenseNet, we directly use the structure of \emph{torchvision}. For STL-10, the training procedure is the same as that of paper \cite{zagoruyko2016wide}, while other hyperparameters, such as learning rate and weight decay, are detailed in the Cutout \cite{devries2017improved}. However, on MNIST and Fashion-MNIST, we train for 50 epochs instead and the scheduler milestones are adjusted proportionally to the 15th, 30th, and 40th epochs. We take the average value of 3 runs for each experimental result. The experiments did not use any image preprocessing or standard data augmentation techniques, such as resize, mirror, flip, and crop. At the same time, we take the parameter $ Ws = 0.8 $ of hidden-level Feature Weaken. As shown in Table \ref{table1}, the hidden-level Feature Weaken method can substantially improve the performance of the model regardless of the deeper models or the weaker models for different model structures. Especially, in testing on STL-10 data, Feature Weaken has a significant improvement effect on different models.
% We use the same training procedure as specified in <<S. Zagoruyko and N. Komodakis. Wide residual networks.British Machine Vision Conference (BMVC), 2016.>>(Cutout也用的是这篇文章里面的参数) on datasets CIFAR-10, CIFAR-100 and STL-10. We train for 200 epochs with batches of 128 images using SGD, Nesterov momentum of 0.9, and weight decay of 5e-4. The learning rate is initially set to 0.1, but is scheduled to decrease by a factor of 5x after each of the 60th, 120th, and 160th epochs.On datasets MNIST and FashionMNIST, we train for 50 instead and the scheduler milestones are adjusted proportionally to the 15th, 30th, and 40th epochs, while the rest of the parameters stay the same. The FeatureWeaken method is applied between the last hidden layer and the classifier layer in all experients, with the default WR as 0.80.

\paragraph{Comparison against other methods:}

% To evaluate the performance of Feature Weaken, we conducted two groups of experiments. One group is based on ResNet-18, and its parameter settings are consistent with the above settings. 
 We use ResNet-18 as the baseline model, and its parameter settings are consistent with the above settings. Experiments are conducted on MNIST, CIFAR-10, CIFAR-100, and STL-10. We report the mean optimal accuracy of 3 runs. The experimental parameters of CIFAR datasets are completely consistent with the above STL-10. The experimental data are evaluated by Top-1 accuracy and Top-5 accuracy. As shown in Table \ref{table2}, we compare it with several methods, Dropout, Mixup, Cutout,  CutMix, and Attentive CutMix. The Dropout parameter is set as 0.5, the $ \alpha $ parameter of Mixup is set to 0.4, the $\alpha$ of CutMix is set to  1, and the $ Path length $ of Cutout is set as 16. These parameter settings have better performance according to the original papers. For Feature Weaken, we still choose the Feature Weaken ($ Ws = 0.8$) on the hidden-level.

According to the experimental results, ResNet-18 combined with Feature Weaken achieves a good improvement. Compared with the baseline, the Top-1 accuracy of CIFAR-10 is improved by 1.24\%, and the Top-5 accuracy is improved by 0.17\%. For CIFAR-100, the increases are 2.70\% and 2.29\%, respectively. Although the experimental results do not exceed CutMix, Feature Weaken can be used with other methods simultaneously. Therefore, according to the experimental results, Feature Weaken can play a positive role in Mixup, CutMix, and Cutout. Especially when Feature Weaken is used with Mixup or CutMix, the advantages of spatial variation of Feature Weaken can be brought into play, and the model can also be encouraged to behave linearly in-between sample categories. When it is used with CutMix, the experimental performance is the best, and the Top-1 accuracy is 3.55\%, 6.95\%, and 14.17\% higher than the baseline in CIFAR-10, CIFAR-100, STL-10.

% In another group of experiments, we use WideResNet-28-10 \cite{zagoruyko2016wide} as a baseline. As shown in Table \ref{tablea}, to clearly show the comparison between FW and other advanced methods, we cited some error rate results of CIFAR-10 from the PatchUp paper \cite{faramarzi2022patchup}. We selected the hard PatchUp parameters, which are the same with the original paper. Obviously, FW is the best result when used with PatchUp.
\paragraph{The Stability of Feature Weaken for CIFAR Classification:}

\begin{figure}[t]
	\centering
	\includegraphics[width=0.9\linewidth]{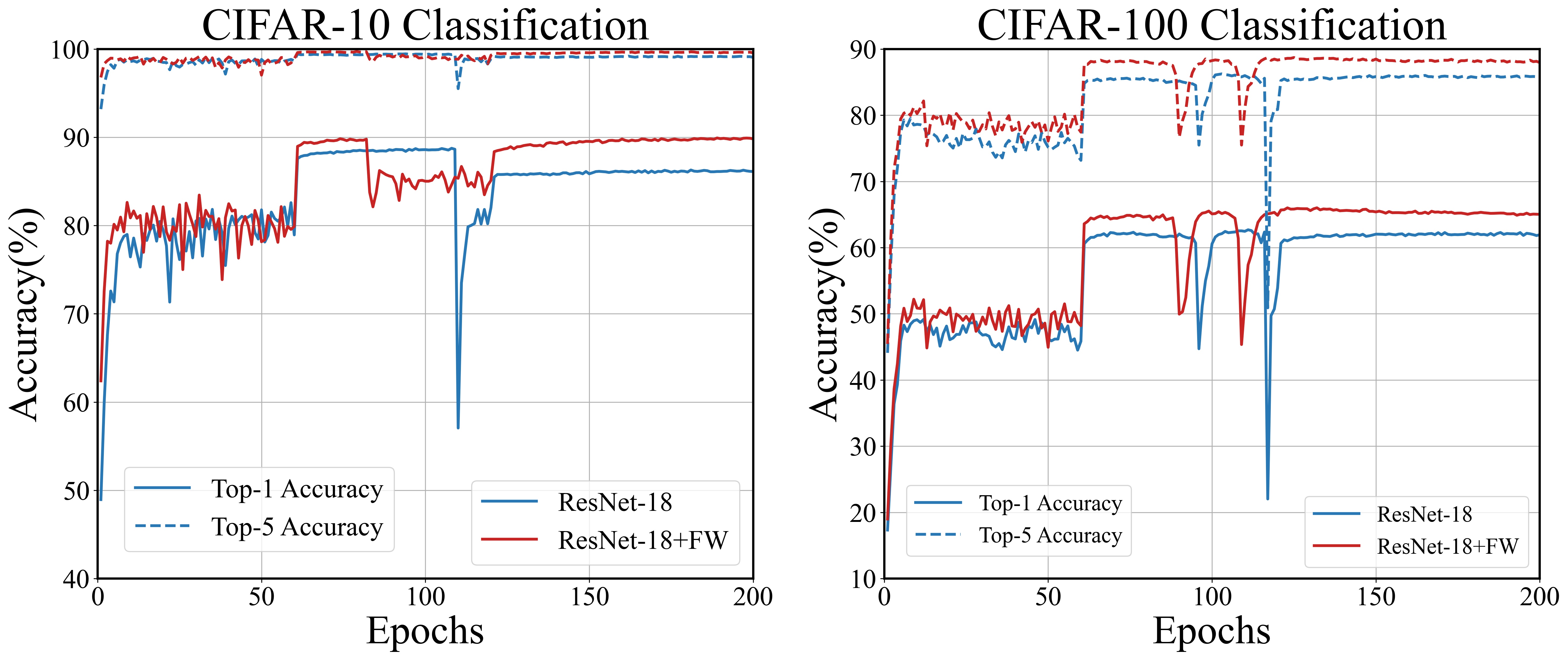}
	\caption{The Top-1(solid) and Top-5(dashed) test accuracy plot of baseline(blue) and Feature Weaken(red) for CIFAR classification.}
	\label{fig3}
\end{figure}
To analyze the influence of Feature Weaken on model iteration, ResNet-18 is selected as the baseline model, two datasets of CIFAR-10 and CIFAR-100 are selected for evaluation, and their test sets are selected as validation sets. The plots of the accuracy of the validation sets with 200 epochs of CIFAR-10 and CIFAR-100 are constructed, as shown in Figure \ref{fig3}.

We can clearly find that when the ResNet-18 with Feature Weaken can not only significantly improve the model performance but also accelerate the convergence. In particular, with the decrease in learning rate at the 120th epoch, the baseline has obvious jitter and decreases the accuracy, but Feature Weaken is more stable.

\paragraph{Feature Weaken with standard data augmentation:}
\begin{table}[t]

	\centering
	\begin{tabular}{ccccc}
		\hline
		%\multicolumn{2}{c}{Part}                   \\
		%\chline
		\multirow{2}{*}{Model}    & \multicolumn{2}{c}{CIFAR-10+}  & \multicolumn{2}{c}{CIFAR-100+}  \\
		\cmidrule(r){2-3} \cmidrule(r){4-5}
		& Top-1&Top-5&Top-1&Top-5\\
		\hline
		ResNet-18  &  94.09&	99.88&	75.67&	93.63
		         \\
		+ Dropout  &   93.92&	99.88&	75.72&	93.83\\
		+ FW-hl &  \textbf{94.44}&	\textbf{99.92}&	\textbf{76.05}&	\textbf{94.08}
		         \\
%		+ droupout +FW &   &   &         \\
		\hline
	\end{tabular}
	\caption{The accuracy(\%) of models with standard data augmentation methods. “+” indicates basic data augmentation (Crop + Horizontal Flip).}
	\label{table3}
\end{table}
Also, to evaluate the validity of Feature Weaken for models using standard data augmentation methods. Consistent with Cutout \cite{devries2017improved}, we used two augmentation methods, random crop and random horizontal flip, on the CIFAR-10 and CIFAR-100 datasets. We take ResNet-18 as the baseline, conduct experiments with Dropout (0.5) and Feature Weaken ($ Ws=0.8 $) respectively, and take the average of the highest accuracy of the 3 runs. The experimental results are shown in Table \ref{table3}. It can be observed that using Feature Weaken can effectively improve the performance compared with baseline or Dropout. The Top-1 accuracy of CIFAR-10 and CIFAR-100 reached 94.44\% and 76.05\%, respectively.

\subsection{Text Classification Experiments}
\begin{table}[t]
	
	\centering
	\begin{tabular}{cccc}
		\hline
		%\multicolumn{2}{c}{Part}                   \\
		%\chline
		Dataset  &Task &Label  & Train/Test Size  \\
		\hline
		SST-2 & sentiment & 2 & 6920/1821    \\
		SST-1 & sentiment &5 & 8544/2210     \\
		TREC-coarse & classification & 6 & 5500/500     \\
		TREC-fine & classification & 47 & 5500/500    \\
		\hline
	\end{tabular}
	\caption{The detail of text datasets.}
	\label{table4}
\end{table}
\paragraph{Text datasets:}As listed in Table \ref{table4}, to evaluate the effect of Feature Weaken for text classification, we perform experiments on four text datasets, SST-1, SST-2, TREC-coarse, and TREC-fine. There are two 2-category datasets and three multi-category datasets. SST are classical sentiment classification datasets, while TREC datasets are sentence classification datasets.

\paragraph{Comparison against other methods:}
We use the Bert model as the backbone model. We use the bert-base-uncased pre-trained model from Huggingface Hub\footnote{\url{https://huggingface.co/bert-base-uncased}} among all experiments. Experiments are initialized by seed 0$\sim$4 and calculate the average result of 5 runs. The Dropout parameter is 0.1, and the maximum length and the batch size of the input sequence are 128 and 32 on the all datasets. We use the AdamW optimizer with a learning rate of 2e-5, eps of 1e-8, weight decay of 1e-4, and epoch of 10. We select three excellent text mixing data augmentation methods: SenMixup\cite{zhang2017mixup,guo2019augmenting}, TMix \cite{chen2020mixtext}, and SSMIX \cite{yoon2021ssmix} to be compared. TMix and SSMIX are the reproduction results of the predecessors' work \cite{yoon2021ssmix}. We set $ \alpha=0.2 $ for SenMixup and TMix. We set window size=10 for SSMIX. For TMix, we set other parameters the same with \cite{yoon2021ssmix}. The specific experimental results are shown in Table \ref{table5}. We can observe that Feature Weaken is also effective for the text classification tasks, and achieve leading results compared with advanced mix data augmentation methods.
\begin{table*}[ht]
	%\caption{The detail of DATA Sets}

	\centering
	\begin{tabular}{ccccc}
		\hline
		%\multicolumn{2}{c}{Part}                   \\
		%\chline
		Model      & SST-1  & SST-2   & TREC-coarse & TREC-fine  \\
		\hline
		Bert  & 54.37&	91.82&	 97.08*&	 86.68*	     \\
		+Dropout (0.5)& 54.14&	92.02&	97.48& 92.48\\
		
		+SenMixup  & 54.30&	92.25&	97.44&	92.08   \\
		+TMix  & 54.13&	92.18&	97.52*&	90.16*    \\
		+SSMIX & 54.33   & 92.03    & 97.60*  & 90.24*  \\
		\hline
		+FW-el ($ Ws=0.2 $) &	54.39&	92.10&	\textbf{97.68}&	92.24 \\
		+FW-el ($ Ws=0.5 $) &	\textbf{54.66}&	92.19&	97.08&	\textbf{92.96}	\\
		+FW-hl ($ Ws=0.8 $) & 	53.99&  92.20&  97.60& 	87.64  \\
		+FW-hl ($ Ws=0.9 $) & 	54.14&  \textbf{92.39}&  97.24& 	82.68   \\
		\hline
	\end{tabular}
	\caption{Performance (accuracy(\%)) of the model with Feature Weaken on text classification tasks. We report the mean accuracy of 5 runs, and the best results are highlighted in bold. We include results from \cite{yoon2021ssmix}*.}
	\label{table5}
\end{table*}

\subsection{Robustness Experiments}
\begin{table*}[t]
	\centering
	\begin{tabular}{ccccc}
		\hline
		%\multicolumn{2}{c}{Part}                   \\
		%\chline
		\multirow{2}{*}{Model}    & \multicolumn{2}{c}{FGSM}& \multicolumn{2}{c}{I-FGSM} \\
		\cmidrule(r){2-3} \cmidrule(r){4-5}
		    & White-box  & Black-box & White-box  & Black-box\\
		\hline
		ResNet-18  & 10.54  & 10.54 & 11.41 & 11.41       \\
		+ Feature Weaken &  33.42&   37.85 & 34.42 & 40.24       \\
		+ Mixup  &   35.14&  52.46  & 36.06 &54.41    \\
		+ Feature Weaken + Mixup  &   27.20&   \textbf{56.57 } & 28.36 & \textbf{58.69}    \\
		+ Cutout &   22.38&    43.81 & 23.35 & 46.46    \\
		+ Feature Weaken + Cutout &   37.46&   46.99  & 38.89 & 49.79    \\
		+ CutMix &   51.98&    49.47& 52.91 & 52.33     \\
		+ Feature Weaken + CutMix &   42.51&   52.39& 43.75 & 55.55     \\
		\hline
	\end{tabular}

	\caption{The results (accuracy(\%)) of Feature Weaken for adversarial samples. The best results (accuracy(\%)) are highlighted in bold.}
	\label{table6}
\end{table*}
Model robustness is a research hotspot of deep learning models \cite{goodfellow2014explaining,su2018robustness,zhang2019theoretically,rebuffi2021data}. Many scholars have proved that deep models are vulnerable to adversarial attacks caused by subtle perturbations \cite{goodfellow2014explaining}. Furthermore, it has been established that the accuracy of the deep model is not related to the robustness of the model \cite{su2018robustness,zhang2019theoretically}. Recently, many scholars have found that data augmentation methods, such as Mixup and CutMix, can well resist adversarial attacks and improve the robustness of models \cite{zhang2017mixup,yun2019cutmix,rebuffi2021data}.

Considering that Feature Weaken can change the sample space vector and the classification boundary of the training samples, we believe Feature Weaken has a certain resistance to anti-attack. To evaluate the influence of Feature Weaken on the model robustness, we use FGSM ($ \epsilon =0.1 $) and I-FGSM ($ \epsilon =0.1$, $ iter=10 $) to add adversarial disturbances to the samples, to generate adversarial samples, and then conduct white-box and black-box tests. In the black-box test, FGSM and I-FGSM are first used to generate adversarial samples for ResNet-18, and then the adversarial samples are used to evaluate the results of other methods. The models used in the experiment are the trained models of CIFAR-10 in Section 4.1.

As shown in Table \ref{table6}, we can observe that Feature Weaken can improve the model robustness and resist adversarial sample attacks in both black-box and white-box tests. Especially, when CutMix, Mixup, and Cutout are combined with Feature Weaken, they can also produce positive gains in model robustness of black-box, and Mixup with Feature Weaken has the highest robustness in the black-box test.
\begin{figure}[t]
	\centering
	\includegraphics[width=0.9\linewidth]{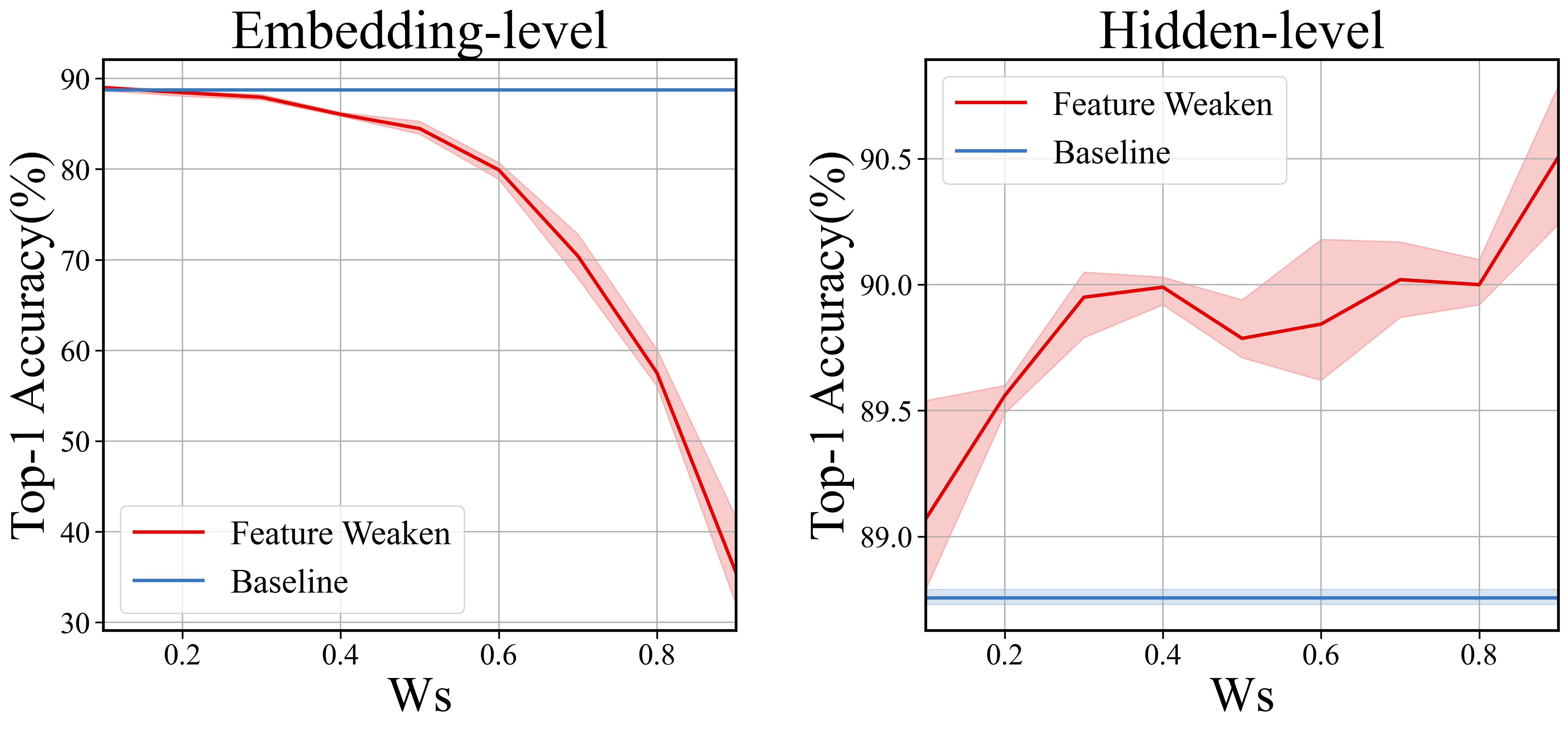}
	\caption{The Weaken Strength ($Ws$) affects the test accuracy plot  (average of three runs, 95\% confidence intervals) for CIFAR-10 classification.}
	\label{fig4}
\end{figure}
\begin{figure*}[!ht]
	\centering
	\subfigure[Original]{
		\includegraphics[width=0.15\linewidth]{dog.png}}
	\subfigure[$ Ws=0.2 $]{
		\includegraphics[width=0.15\linewidth]{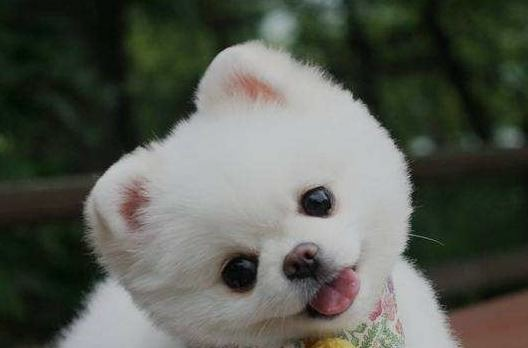}}
	\subfigure[$ Ws=0.5 $]{
		\includegraphics[width=0.15\linewidth]{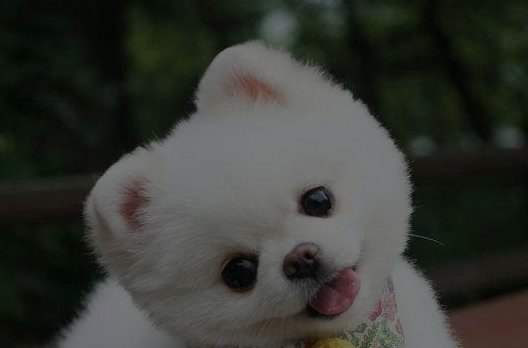}}
	\subfigure[$ Ws=0.8 $]{
		\includegraphics[width=0.15\linewidth]{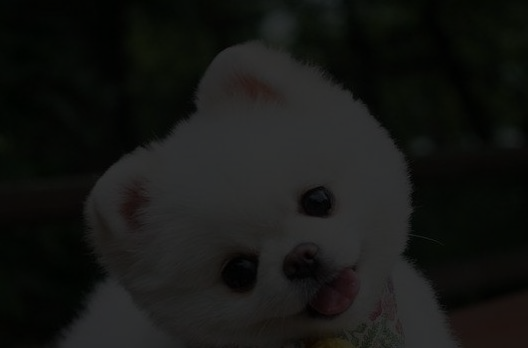}}
	\subfigure[$ Ws=0.9 $]{
		\includegraphics[width=0.15\linewidth]{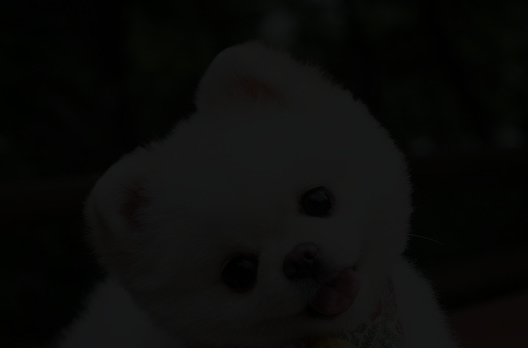}}
	\subfigure[$ Ws=0.99 $]{
		\includegraphics[width=0.15\linewidth]{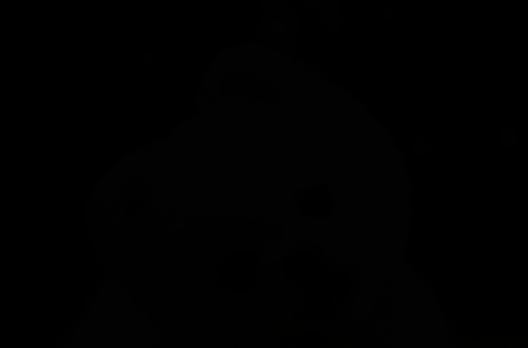}}
	\caption{Feature Weaken on the embedding-level of the example images.}
	\label{Fig5}
\end{figure*}
\subsection{Ablation Studies}

We perform ablation experiments on the CIFAR-10 dataset and evaluate a set of $ Ws $ parameters for FW-el and FW-hl. As shown in Figure \ref{fig4}, for hiddle-level, the $Ws$ is approximately monotonically increasing between 0.1 and 0.9. With the increase of $Ws$, the accuracy is getting higher. While the embedding-level shows a monotonically decreasing effect with the increase of $Ws$. Overall, the effect of hidden-level Feature Weaken is generally better than embedding-level Feature Weaken. The optimal parameter is when $ Ws $ is 0.9 in hidden-level, the accuracy of the test set reaches 90.51\%.

To observe the changes in weakened sample space, we perform an embedding-level Feature Weaken operation  and show the weakened sample of different $Ws$. As shown in Figure \ref{Fig5}, as the weaken strength increases, the color of the image has darkened and the texture becomes relatively blurry. When $ Ws $ is 0.99, the image is approximately black.

\section{Conclusion}

We propose a new regularization training method, Feature Weaken, which is different from the existing Feature Drop, Feature Mix, and
other regularization methods. We consider the weakening of data features on the embedding-level and the hidden-level to generate the vicinal data distribution of the original samples. We then use the weakened features for model training. Feature Weaken can be regarded as a data augmentation method. Through experiments, we prove that Feature Weaken has good universality and can be used with other methods to produce significant improvement. And in the image or text classification task, compared with the feature mixing or feature dropping model, it has achieved a leading performance. We also analyze and discuss the function of Feature Weaken. On the one hand, Feature Weaken operation changes the process of loss optimization. Compared with the original method, its gradients are  reduced during back-propagation, which is similar to the function of learning rate or weight decay, to increase the difficulty of model training. On the other hand, the spatial position of the original training sample changed by Feature Weaken does not change the spatial angle of the sample vector. Feature Weaken makes the classification boundaries of the samples more difficult to distinguish, and the training is more robust.

In the future, we will continue to study and explore the application scope of Feature Weaken, enrich its functions, and extend from global mode to local mode.

% References and End of Paper
% These lines must be placed at the end of your paper
\fontsize{9.5pt}{10.5pt} \selectfont

\bibliography{ref.bib}

% \section{Acknowledgments}

% \bigskip
% \noindent Thank you for reading these instructions carefully. We look forward to receiving your electronic files!

\end{document}